\documentclass{article}
\usepackage{spconf,amsmath,graphicx,mathtools}
\usepackage[utf8]{inputenc}

\def\q{{\mathbf q}}
\def\x{{\mathbf x}}

\title{VOICE TRIGGER DETECTION FROM LVCSR HYPOTHESIS LATTICES USING BIDIRECTIONAL LATTICE RECURRENT NEURAL NETWORKS}
\name{Woojay Jeon, Leo Liu, and Henry Mason}
\address{Apple\\ One Apple Park Way, Cupertino, California\\ \small \tt{\{woojay,lliu9,hmason\}@apple.com}}
\begin{document}
%
\maketitle
\begin{abstract}
We propose a method to reduce false voice triggers of a speech-enabled personal assistant by post-processing the hypothesis lattice of a server-side large-vocabulary continuous speech recognizer (LVCSR) via a neural network.
We first discuss how an estimate of the posterior probability of the trigger phrase can be obtained from the hypothesis lattice using known techniques to perform detection, then investigate a statistical model that processes the lattice in a more explicitly data-driven, discriminative manner.
We propose using a Bidirectional Lattice Recurrent Neural Network (LatticeRNN) for the task, and show that it can significantly improve detection accuracy over using the 1-best result or the posterior.
\end{abstract}
\section{Introduction}
\label{sec:intro}

Speech-enabled personal assistants are often conveniently activated by use of a trigger phrase.
In the case of the Apple personal assistant Siri, English users can say ``Hey Siri'' to activate the assistant and make a request in a single step, e.g. ``Hey Siri, how is the weather today in Cupertino?''

Typically, an on-device detector \cite{Sigtia2018} decides whether the trigger phrase was spoken, and if so allows the audio (including the trigger phrase) to flow to a server-based large vocabulary continuous speech recognizer (LVCSR).
Because the on-device detector is resource-constrained, its accuracy is limited and leads to occasional ``false triggers'' where the user did not speak the trigger phrase but the device wakes up anyway and gives an unexpected response.

To reduce false alarms, one could conceive of a secondary trigger phrase detector running on the server, utilizing a much larger statistical model than the one on the device to more accurately analyze the audio and override the device's trigger decision when it sees that no trigger phrase is present.
This method would optimally improve the accuracy because we are using a dedicated acoustic model that is specifically trained for the detection task.
However, since this must be done for every utterance, a more resource-efficient approach is to use the output of the server-side LVCSR, which is run anyway for every utterance, to perform the secondary detection.

The obvious method is to check whether the top recognition result produced by the LVCSR begins with the trigger phrase or not.
However, the LVCSR is often biased toward recognizing the trigger phrase at the beginning of the audio, and therefore ``hallucinates'' the phrase in the 1-best result even though it does not exist.

LVCSR output has been leveraged to keyword spotting (or keyword search, which is closely related) in many past studies. An early method used the sum of the likelihoods of hypotheses containing the keyword in an $n$-best list \cite{weintraub1995}. However, an $n$-best list is a lossy representation of the word hypothesis lattice, which is a richer representation of the output of an ASR \cite{saraclar2004}. Hence, subsequent works have acted directly on the hypothesis lattice by computing word posteriors \cite{szoke2005} or normalized confidence scores \cite{karakos2013} or contextual features via neural networks \cite{chen2017}.

We will begin by examining how a secondary voice trigger detector can be built using known processing techniques on the LVCSR's hypothesis lattice to compute a posterior probability for the trigger phrase \cite{szoke2005}.
This method, however, is strictly limited by the reliability of the acoustic and language models of the LVCSR, which may not be accurate enough for falsely-triggered audio that often contains diverse and unpredictable sounds that may or may not be speech.
To actively overcome some of the LVCSR's errors, we consider the use of a statistical model that can interpret the hypothesis lattice in a discriminative, data-driven manner.
We propose the use of a bidirectional version of ``LatticeRNN'' \cite{ladhak2016} for this purpose, and show that a significant gain in accuracy can be obtained compared to using the simple posterior probability.

\section{Voice trigger detection based on lattice posteriors}
\label{sec:posteriors}

Consider the probability that a speech utterance with acoustic features $X$ starts with a designated trigger phrase.
The trigger phrase is a fixed sequence of $n$ words, $V= [ v_1, v_2, \cdots, $ $v_n ]$.
The probability of the first $n$ words of the utterance, $w_1, \cdots, w_n$ being the trigger phrase is
\begin{equation}\label{post}
	P\left( {\left. {{w_1} = {v_1},{w_2} = {v_2}, \cdots ,{w_n} = {v_n}} \right|X} \right)
\end{equation}

If we can compute the above probability, we could simply apply a threshold to obtain a trigger phrase detector.

The probability in Eq. \eqref{post} can be written as
\begin{equation}
\begin{spreadlines}{-1em}
\begin{aligned}\label{post2}
\sum\limits_{{r_1},{r_2} \cdots }^{} P( & {w_1} = {v_1},, \cdots ,{w_n} = {v_n},  \\
& \left. {{w_{n + 1}} = {r_1},{w_{n + 2}} = {r_2}, \cdots } \right|X)
\end{aligned}
\end{spreadlines}
\end{equation}
where $r_1, r_2, \cdots$ are the words following the trigger phrase.

With some abuse of notation, we write this as
\begin{equation}\label{post3}
\sum\limits_R^{} {P\left( {\left. {V,R} \right|X} \right)}  = \frac{{\sum\limits_R^{} {p\left( {V,R,X} \right)} }}{{\sum\limits_W^{} {p\left( {W,X} \right)} }}
\end{equation}
where $V$, $R$, and $W$ represent the trigger phrase terms, the words following the trigger phrase, and all the words in the utterance, respectively.

If we assumed that the hypothesis lattice provided by an LVCSR spans the entire space of all possible word sequences for $R$ and $W$ (which is obviously a significant approximation, since the lattice would be heavily pruned and show only a small set of possible hypotheses), Eq. \eqref{post3} can be solved in a straightforward way using the well-known lattice forward-backward algorithm \cite{szoke2005, wessel01}.

Stated more formally for our specific case, we have
\begin{equation}\label{post4}
p\left( {V,R,X} \right) = \sum\limits_{\mathbf{q}}^{} {p\left( {V,R,{\mathbf{q}},X} \right)}  = \sum\limits_{{\mathbf{q}} \in {Q_{VR}}}^{} {p\left( {X,{\mathbf{q}}} \right)}
\end{equation}
where $\q$ is a path through the hypothesis lattice, and $Q_{VR}$ is the set of all paths that contain the word sequence $V+R$.

Let $\q_V$ be the front part of $\q$ that contains the trigger phrase $V$ and $\q_R$ be the remainder of $\q$ that contains the rest of the words $R$.
Also let $X_V$ and $X_R$ denote the speech frames consumed by $\q_V$ and $\q_R$, respectively. We now have
\begin{equation}\label{post5}
p\left( {V,R,X} \right) = \sum\limits_{{{\mathbf{q}}_V} \in {Q_V}}^{} {\sum\limits_{{{\mathbf{q}}_R} \in {Q_R}\left( {{{\mathbf{q}}_V}} \right)}^{} {p\left( {{X_V},{X_R},{{\mathbf{q}}_V},{{\mathbf{q}}_R}} \right)} }
\end{equation}
where $Q_V$ is the set of all initial partial paths that contain $V$, and $Q_R(\q_V)$ is the set of all residual paths that follow $\q_V$.

This becomes
\begin{equation}\label{post6}
p\left( {V,R,X} \right) = \sum\limits_{{{\mathbf{q}}_V} \in {Q_V}}^{} {p\left( {{X_V},{{\mathbf{q}}_V}} \right)\beta \left( {{X_R},{{\mathbf{q}}_v}} \right)}
\end{equation}
where 
\begin{equation}\label{backward}
\beta \left( {{X_R},{{\mathbf{q}}_v}} \right) = \sum\limits_{{{\mathbf{q}}_R} \in {Q_R}\left( {{{\mathbf{q}}_V}} \right)}^{} {p\left( {{X_R},{{\mathbf{q}}_R}} \right)}
\end{equation}

The joint distribution $p(X_V, \q_V)$ in Eq. \eqref{post6} is simply obtained by multiplying the joint probabilities along the path $\q_V$.
If $\q_V$ consists of $n$ arcs $\lambda_1, \cdots, \lambda_n$, each $i$'th arc consuming the acoustic features $X_i$ and storing an acoustic model score ${p\left( {\left. {{X_i}} \right|{\lambda _i}} \right)}$ and a contextual transition score (which includes the language model score and pronunciation score) ${P\left( {\left. {{\lambda _i}} \right|{\lambda _1} \cdots {\lambda _{i - 1}}} \right)}$, we have
\begin{equation}\label{arc_score}
p\left( {{X_V},{{\mathbf{q}}_V}} \right) = \prod\limits_{i = 1}^n {p\left( {\left. {{X_i}} \right|{\lambda _i}} \right)P\left( {\left. {{\lambda _i}} \right|{\lambda _1} \cdots {\lambda _{i - 1}}} \right)}
\end{equation}

$\beta (X_R, \q_V)$ in Eq. \eqref{post6} is the ``backward'' score of the node at the end of path $\q_V$ that we obtain by the lattice forward-backward algorithm.

\section{Bidirectional Lattice-RNN for voice trigger detection}
\label{sec:latticernn}

In the previous section, we discussed how to compute a voice trigger posterior probability from the hypothesis lattice to perform voice trigger detection.
A fundamental limitation to such an approach is that it is directly exposed to errors in the LVCSR's acoustic and language model scores.
If the LVCSR is overly biased toward giving high scores to the voice trigger phrase, the posterior in Eq. \eqref{post3} will be consistently high and the detection accuracy will suffer.
The only tunable parameter in the system is the detection threshold applied to the posterior, and one parameter (applied across all utterances) is not sufficient for overcoming modeling errors in the LVCSR.

This motivates us to build a more general statistical model with many more parameters that can use training examples to learn how to process the hypothesis lattice in a data-driven manner.
In effect, the model learns the ``mistakes'' in the LVCSR's scores and actively tries to correct them. 
In our proposed method, we employ Lattice Recurrent Neural Networks \cite{ladhak2016} that can read entire hypothesis lattices without requiring us to heuristically convert them into lossy forms such as $n$-best lists or word confusion networks.

\subsection{The Bidirectional LatticeRNN}
\label{subsec:bidir}

``LatticeRNN'' \cite{ladhak2016} was originally introduced for the task of classifying user intents in natural language processing.
For a topologically-sorted hypothesis lattice, the feature vector of each arc is input to the neural network along with the state vector of the arc's source node, and the output of the neural network becomes the arc's state.
For a given arc $e$ that has input feature vector $\x(e)$ whose source node $p(e)$ possesses state vector ${{\mathbf{h}_f}\left( {p\left( e \right)} \right)}$, the neural network characterized by input transformation $U_f$, state transformation $V_f$, bias $\mathbf{b}_f$, and activation function $g(\cdot)$ outputs the arc's state $\mathbf{h}_f(e)$:
\begin{equation}\label{forward_rnn}
{{\mathbf{h}}_f}\left( e \right) = g\left\{ {U_f^T{\mathbf{x}}\left( e \right) + V_f^T{{\mathbf{h}}_f}\left( {p\left( e \right)} \right) + {{\mathbf{b}}_f}} \right\}
\end{equation}

For a given node $s$, the state vector ${\mathbf{h}_f}\left( s \right)$ is obtained via a pooling function applied to the states of all incoming arcs:
\begin{equation}\label{forward_pool}
{{\mathbf{h}}_f}\left( s \right) = {f_{pool}}\left( {\left\{ {{{\mathbf{h}}_f}\left( e \right):n\left( e \right) = s} \right\}} \right)
\end{equation}

Since information propagates only in a forward direction in this neural network, the state of an arc is determined only by the arcs that precede it, and is unaffected by any arc that succeeds it.
If we imagined the state of the arcs for the words ``hey'' and ``Siri'' as a measure of how relevant they are for the detection task (analogous to their posterior probabilities), it would be desirable for the states to also depend on the succeeding words. Words asking about the weather, for instance, should make the trigger word arcs more relevant than a random string of words resembling a TV commercial.
Based on this intuition, we also perform the same sort of propagation in the reverse direction, in a similar manner as is done for conventional RNNs \cite{schuster1997}.

In parallel to the forward propagation in Eq. \eqref{forward_rnn}, we have another neural network parameterized by $U_b$, $V_b$, and $\mathbf{b}_b$, which takes arc $e$'s feature vector $\x(e)$ and its destination node's \emph{backward} state vector ${{{\mathbf{h}}_b}\left( {n\left( e \right)} \right)}$ and outputs the arc's backward state vector ${{\mathbf{h}}_b}\left( e \right)$:
\begin{equation}\label{backward_rnn}
{{\mathbf{h}}_b}\left( e \right) = g\left\{ {U_b^T{\mathbf{x}}\left( e \right) + V_b^T{{\mathbf{h}}_b}\left( {n\left( e \right)} \right) + {{\mathbf{b}}_b}} \right\}
\end{equation}

The backward state vector for a node $s$ is obtained by applying the pooling function to the outgoing arcs:
\begin{equation}\label{backward_pool}
{{\mathbf{h}}_b}\left( s \right) = {f_{pool}}\left( {\left\{ {{{\mathbf{h}}_b}\left( e \right):p\left( e \right) = s} \right\}} \right)
\end{equation}

The forward state vector ${{\mathbf{h}}_f \left( {{s_{\rm term}}} \right)}$ of the lattice's terminal node and the backward state vector ${{\mathbf{h}}_b \left( {{s_{\rm init}}} \right)}$ of the lattice's initial node are concatenated to form a single state vector ${{\mathbf{h}}_{\rm lat}}$  that represents the entire lattice.

\begin{equation}\label{lattice_state}
{{\mathbf{h}}_{\rm lat}} = {\left[ {{\mathbf{h}}_f^T\left( {{s_{\rm term}}} \right) \quad {\mathbf{h}}_b^T\left( {{s_{\rm init}}} \right)} \right]^T}
\end{equation}

An additional feed forward network is then applied to ${{\mathbf{h}}_{\rm lat}}$ to output a single scalar value that represents how likely the input lattice starts with the voice trigger phrase.

\section{EXPERIMENT}
\label{sec:experiment}

\begin{table}[htb]
\centering
\begin{tabular}{ l | c | c | c }
 Data Type & VT  & No VT & Total \\
\hline
  Training    & 12,271 & 6,731  & 19,002 \\
  Development & 3,347  & 1,836  & 5,183 \\
  Evaluation  & 6,693  & 3,672  & 10,365 \\
\end{tabular}
\caption[]{Counts of utterances with (VT) and without (no-VT) the voice trigger phrase collected for this experiment. Note that ``no-VT'' utterances are relatively rare in production because most of them are immediately discarded by the device and it is only those that are falsely accepted by the device-side detector that can become part of our data. Hence, the ``no-VT'' utterances must be collected over a much longer period of time than ``VT'' utterances.}
\label{tab:counts}
\end{table}

\begin{table}[htb]
\centering
\begin{tabular}{ l | c | c | c | c }
 Method    & $P_M$ (\%) & $P_{FA}$ (\%) & EER (\%) \\
\hline
  Baseline      & 0.84 & 78.92 & - \\
  Posterior     & 0.84 & 79.08 & 35.68 \\
  Lattice RNN   & 0.84 & 22.55 & 5.51 \\
  Bidir Lat RNN & 0.84 & 17.05 & 4.59 \\
\end{tabular}
\caption[]{Results on development data, showing the Probability of Miss ($P_M$), Probability of False Alarm ($P_{FA}$), and Equal Error Rate (EER). The ``Baseline'' method is to check the 1-best result of the LVCSR, and is not tunable. The ``Posterior'' method is to threshold the trigger phrase posterior probability in Section \ref{sec:posteriors}. ``Lattice RNN'' and ``Bidir Lat RNN'' are the neural network models in Section \ref{sec:latticernn}. For each of the bottom three methods, $P_M$ and $P_{FA}$ is from the operating point where $P_M$ is closest to the baseline $P_M$.}
\label{tab:result-cv}
\end{table}

\begin{table}[htb]
\centering
\begin{tabular}{ l | c | c | c }
 Method    & $P_M$ (\%) & $P_{FA}$ (\%) \\
\hline
  Baseline      & 0.84 & 78.46 \\
  Posterior     & 0.84 & 78.87 \\
  Lattice RNN   & 1.20 & 20.81  \\
  Bidir Lat RNN & 1.15 & 17.57  \\
\end{tabular}
\caption[]{Results on evaluation data. For the bottom three methods, detection thresholds corresponding to the operating points in Table \ref{tab:result-cv} (with fixed $P_M$ on the development data) were used to obtain the values in this table.}
\label{tab:result-ev}
\end{table}

A labeled set of utterances was used for the experiment, where some began with the ``Hey Siri'' trigger phrase and the rest did not. Table \ref{tab:counts} shows the number of positive and negative examples used for training, development, and evaluation.

The ``baseline'' method is to simply look at the top recognition result from the LVCSR and check whether it begins with ``Hey Siri'' or not.
As shown in Tables \ref{tab:result-cv} and \ref{tab:result-ev}, the Probability of Miss (failing to recognize the trigger phrase when it is present) is usually less than 1\%, but the Probability of False Alarm (``hallucinating'' the trigger phrase when it is not present) is around 79\% on this data set.
Note that an actual Siri user would experience far less false alarms because most utterances without the trigger phrase get immediately discarded by the device-side detector, and it is only those occasional few that slip past the detector that become part of our negative data.

\begin{figure}[]
  \centering
  \includegraphics[width=2.4in]{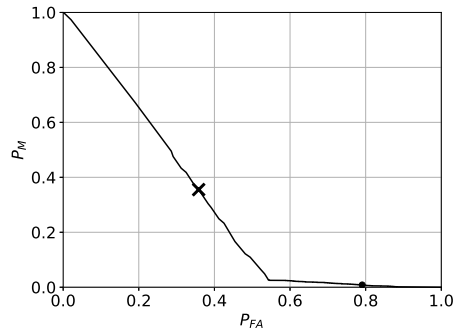}
  \vspace{-.15in}
  \caption{ROC curve for posterior-based voice trigger detection on development data showing Probability of Miss ($P_M$) vs. Probability of False Alarm ($P_{FA}$). ``$\times$'' marks the equal error rate operating point and ``\textbullet'' marks the point where $P_M$ is closest to the baseline's.}
  \label{fig:post_cv_roc}
\end{figure}

The ``posterior'' method is that described in Section \ref{sec:posteriors}, where the voice trigger posterior probability is directly computed from the hypothesis lattice.
As is evident in the ROC curve in Figure \ref{fig:post_cv_roc}, above a certain threshold (where there is a clear sharp angle in the curve) the voice trigger posterior tends to be evenly distributed between true triggers and false triggers and is hence a poor discriminant.
Below the turning point, however, most of the inputs are false triggers, so the detector performs much better when the false alarm probability is around 55\% or higher.
For the development data, the $P_{FA}$ is 79.08\% when $P_M$ is the same as the baseline (0.84\%).
The threshold obtained from this operating point was applied to the evaluation data to obtain the values in Table \ref{tab:result-ev}.

For the lattice RNNs, the arc feature vector $\x(e)$ consists of 19 features: the log acoustic score, the log language model score, the number of speech frames consumed by the arc, a binary feature indicating whether the word is ``hey'',  a binary feature indicating whether the word is ``Siri'', and 14 features representing the phone sequence of the arc's word.
The phone sequence, which is variable length, is converted to a 51-dimensional binary bag-of-phones vector and reduced to 14 dimensions via an autoencoder.
The autoencoder is trained using a lexicon of pronunciations for 700K words.
The unidirectional lattice RNN has 24 dimensions for the state vector, which is fed to a feedforward network with 20 hidden nodes, resulting in a total 1,577 parameters. The bidirectional lattice RNN has 15 dimensions in each state vector, and is used with a feedforward network with 15 hidden nodes, resulting in a total 1,531 parameters. 
All inputs are mean- and variance-normalized, with the scale and bias computed from the training data. The pooling function in Eq. \eqref{forward_pool} and \eqref{backward_pool} is the arithmetic mean.

A huge accuracy gain is observed when using the lattice neural network compared to the baseline or posterior-based method, and more when using the bidirectional instead of unidirectional lattice RNN. 

In terms of runtime computational complexity, the proposed method adds minimal latency to the existing LVCSR because 1) the lattice is usually compact; over the training data, the average number of arcs per lattice is 42.7 whereas the average number of acoustic feature frames is 406, and 2) the lattice RNN is small, with only around 1,500 parameters.

To retain the order of the phone sequence in each arc, we also tried replacing the bag-of-phones features with the encoding from a sequence-to-sequence autoencoder, but did not observe accuracy improvement with the given data.

\section{CONCLUSION AND FUTURE WORK}
\label{sec:conclusion}

We have proposed a novel method of voice trigger phrase detection based on the hypothesis lattice of an LVCSR using a Bidirectional Lattice Recurrent Neural Network, and showed that it can significantly reduce the occurrence of false triggers in a digital personal assistant.

Given that the LVCSR is being used for both trigger detection and speech recognition in this case, future work will investigate an objective function that jointly maximizes both recognition accuracy and detection accuracy, which would better fit the true goal of the system.

\section{ACKNOWLEDGEMENTS}
\label{sec:ack}

Hywel Richards and John Bridle made helpful comments that improved the overall quality of this paper. False trigger data used in experiments was originally collected via efforts coordinated by Srikanth Vishnubhotla.

\begin{figure}[]
  \centering
  \includegraphics[width=3.0in]{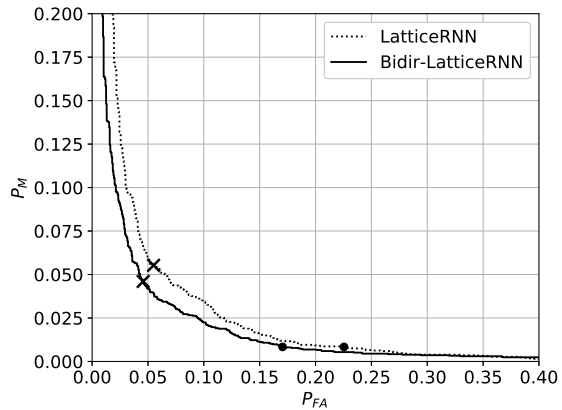}
  \vspace{-.15in}
  \caption{ROC curve on development data using Lattice RNN and Bidirectional Lattice RNN. ``$\times$'' indicates the equal error rate operating points and ``\textbullet'' indicates the operating point where $P_M$ is closest to the baseline's.}
  \label{fig:nnet_cv_roc}
\end{figure}

\begin{figure}[]
  \centering
  \includegraphics[width=3.0in]{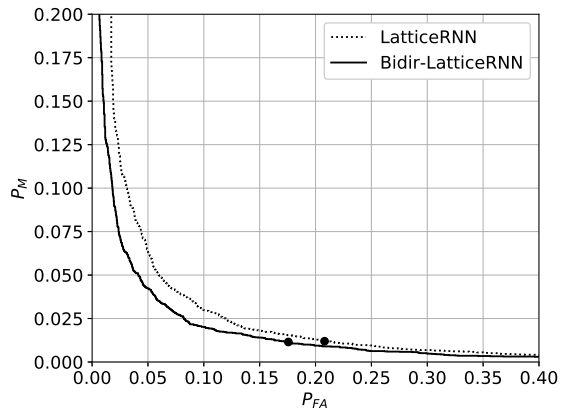}
  \vspace{-.15in}
  \caption{ROC curve on evaluation data for voice trigger detection using Lattice RNN and Bidirectional Lattice RNN.  The ``\textbullet'' for each curve indicates the operating point where the threshold from the ``\textbullet'' in Figure \ref{fig:nnet_cv_roc} is applied to the evaluation data.}
  \label{fig:nnet_ev_roc}
\end{figure}



\bibliographystyle{IEEEbib}
\bibliography{strings,refs}

\end{document}